\documentclass[times, preprint, 10pt]{elsarticle}

\usepackage{amsmath,amssymb,amsthm,mathtools,bm}
\usepackage{graphicx}
\usepackage{booktabs,multirow,array}
\usepackage{adjustbox}
\usepackage{subcaption}
\usepackage{algorithm}
\usepackage{algpseudocode}
\usepackage{enumitem}
\usepackage{xcolor}
\usepackage[hidelinks]{hyperref}
\hypersetup{pdfauthor={Zhenzhi Tang}}
\newtheorem{theorem}{Theorem}
\newtheorem{proposition}{Proposition}

\theoremstyle{definition}

\makeatletter\def\ps@pprintTitle{\ps@plain}\makeatother

\begin{document}

\begin{frontmatter}

\title{Who gets the information? An \texorpdfstring{$\alpha$}{alpha}-fair label-coverage framework for multi-label feature selection}

\author[cmu]{Zhenzhi Tang\corref{cor1}}
\ead{zhenzhit@andrew.cmu.edu}
\cortext[cor1]{Corresponding author.}
\affiliation[cmu]{organization={Information Networking Institute, Carnegie Mellon University},
            city={Pittsburgh},
            state={PA 15213},
            country={USA}}

\begin{abstract}
Recent information-theoretic filters for multi-label feature selection disagree on how to distribute relevance over labels: some amplify features that inform a few labels, others balance what every label receives. We show these choices are implicit social welfare functions. Treating labels as agents whose utility is the information they receive, we propose FLIC, which greedily maximises the $\alpha$-fair welfare of per-label information coverage. A single parameter $\alpha$ moves from utilitarian ($\alpha=0$) through proportionally fair ($\alpha=1$) to max--min selection ($\alpha\to\infty$), and a dozen published relevance terms lie on this axis, most as special cases, first-order approximations or limits. The objective is monotone submodular for every $\alpha\ge0$ whenever each label's information is, so greedy selection is a $(1-1/e)$-approximation; this fails for $\alpha<0$, and the max--min limit is inapproximable. Fano's and the Hellman--Raviv inequalities tie raw coverage to the Bayes Hamming loss and normalised coverage to per-label relative error. FLIC uses bias-corrected, label-wise conditional information and has no relevance--redundancy trade-off parameter. On 22 data sets, FLIC is the best of 14 information-theoretic filters on seven of nine metrics, winning 86 of 117 pairwise comparisons significantly and losing none significantly. Among 17 methods it ranks first on ranking loss, coverage and macro-AUC with two classifiers and has the best mean rank (narrowly with ML-kNN), while three embedded methods lead on Hamming loss, one-error and micro-F1.
\end{abstract}

\begin{keyword}
Multi-label learning \sep Feature selection \sep Information theory \sep $\alpha$-fairness \sep Submodularity \sep Label imbalance
\end{keyword}

\end{frontmatter}

\section{Introduction}\label{sec:intro}

Multi-label learning associates every instance with a set of labels---the topics of a document, the objects in an image, the functions of a gene \cite{zhang2014review}---and benchmark collections such as those distributed with Mulan \cite{tsoumakas2011mulan} span text, images, audio and biology. Multi-label data are typically high dimensional, and irrelevant or redundant features degrade both the accuracy and the cost of the downstream predictor. Feature selection, which keeps a small subset of the original and therefore interpretable features, is the usual remedy \cite{guyon2003}. Information-theoretic filters are particularly popular for multi-label feature selection (MLFS): they are classifier independent and build on the max-relevance min-redundancy principle of mRMR \cite{peng2005}, whose low-order approximations of mutual information were unified by Brown et al.\ \cite{brown2012}. A long line of multi-label filters has refined how relevance to many labels is measured and how redundancy with the selected features is discounted: PMU \cite{lee2013pmu} and D2F \cite{lee2015d2f} approximate high-order interactions between features and labels, MDMR \cite{lin2015mdmr} combines max-dependency with conditional redundancy, SCLS \cite{lee2017scls} scales redundancy by the relevance of the candidate, and LRFS \cite{zhang2019lrfs}, DCR-MFS \cite{zhang2021dcr} and STFS \cite{gao2023stfs} condition relevance on labels and selected features.

This paper is about a question that every such filter answers implicitly but that has not been posed explicitly: \emph{when a feature budget is shared by many labels, which labels should receive the information?} The answers in recent work are inconsistent. Classical criteria such as PMU \cite{lee2013pmu}, D2F \cite{lee2015d2f} and SCLS \cite{lee2017scls} sum the mutual information over labels, so frequent, high-entropy labels dominate the score. Gonz\'alez-L\'opez et al.\ \cite{gonzalez2020} contrast the Euclidean norm of the per-label mutual informations (ENM), which rewards features that are very informative about some labels, with their geometric mean (GMM), which rewards features that inform all labels. LSRIFS \cite{han2025lsrifs} argues that relevance concentrated on few labels should be amplified, whereas RRFS \cite{han2024rrfs} weights labels inversely to the information they have already received, and CIRFS \cite{liu2026cirfs} does the same with the inverse \emph{square root} of the received information, mixed with the label frequency. Each heuristic is motivated by an example, none derives its aggregation over labels from an explicit objective on the selected set, and they recommend opposite behaviours---or, for RRFS and CIRFS, the same behaviour with different strengths.

We argue that these heuristics are \emph{social welfare functions}. Treat each label as an agent and the information it receives from the selected features, $I(X_S;Y_l)$, as its utility; a selection criterion then decides how a common budget is shared among the agents. Network resource allocation offers a canonical one-parameter family for this decision, the $\alpha$-fair utilities of Mo and Walrand \cite{mo2000}: $\alpha=0$ maximises total utility, $\alpha=1$ yields the proportional fairness of Kelly et al.\ \cite{kelly1998}, $\alpha\to\infty$ yields max--min fairness; extended to $\alpha<0$, as we do here, the family rewards concentration. Bertsimas et al.\ \cite{bertsimas2011} quantify the efficiency that fairness costs. Building on this observation, we make the following contributions.

\begin{enumerate}[leftmargin=1.6em,itemsep=1pt]
\item \textbf{Framework.} We formulate MLFS as maximising the $\alpha$-fair welfare of per-label information coverage and show that the relevance terms of a dozen published filters lie on one axis: those of MIM, PMU, D2F, SCLS, MDMR, STFS and DCR-MFS ($\alpha=0$), GMM and RRFS ($\alpha=1$) and ENM ($\alpha=-1$) are special cases, first-order approximations or limits of one objective, while CIRFS interpolates the first-order $\alpha=1/2$ weights with label frequencies and LSRIFS multiplies the $\alpha=-1$ and $\alpha=0$ scores (Section~\ref{sec:framework}).
\item \textbf{Theory.} The objective is monotone submodular for every $\alpha\ge0$ whenever each label's information is, so greedy selection is a $(1-1/e)$-approximation; the guarantee fails for $\alpha<0$, and the max--min limit is inapproximable. Fano's and the Hellman--Raviv inequalities tie raw coverage to the Bayes Hamming loss and normalised coverage to each label's relative error, which suggests that small $\alpha$ favours Hamming loss and larger $\alpha$ label-wise metrics.
\item \textbf{Algorithm.} FLIC (\emph{Fair Label-Information Coverage}) maximises the objective greedily with a chain-rule coverage ledger and label-wise, bias-corrected conditional information; its welfare gains are exact rather than first-order, so if its label-wise gains equalled the true conditional information given the selected features, FLIC would be exactly the greedy algorithm of Theorem~\ref{thm:submod}. It has no relevance--redundancy trade-off parameter and runs in $O(knqd)$ time (Section~\ref{sec:algorithm}).
\item \textbf{Evidence.} On 22 data sets, with all design choices fixed on separate development data, FLIC is the best of 14 information-theoretic filters on seven of nine metrics, with 86 of 117 pairwise comparisons significantly in its favour and none against, and it remains the best filter when all filters use its bias correction (a control run with ML-kNN). Among all 17 compared methods, including three tuned embedded methods, it ranks first on ranking loss, coverage and macro-AUC and has the best mean rank with both classifiers---clearly with logistic regression, narrowly with ML-kNN---although the embedded methods lead on Hamming loss, one-error and micro-F1 (with ML-kNN also on average precision); its mean AUC on rare labels exceeds that of 15 of the 16 baselines (Section~\ref{sec:results}). In a controlled study and, on the raw coverage scale, over the benchmarks, increasing $\alpha$ to a moderate value ($2$ to $4$) trades Hamming loss and micro-F1 for macro-AUC and Tail-AUC, as the theory suggests.
\end{enumerate}

\section{Related work}\label{sec:related}

\paragraph{Information-theoretic multi-label feature selection}
Algorithm-adaptation filters score features against all labels directly. PMU \cite{lee2013pmu} approximates the joint information between the features and all labels by a truncated M\"obius expansion; D2F \cite{lee2015d2f} keeps the feature--feature--label interactions, SCLS \cite{lee2017scls} scales redundancy by the relevance of the candidate, and MDMR \cite{lin2015mdmr} combines max-dependency with conditional redundancy. Later criteria refined the relevance term with conditional mutual information: LRFS distinguishes dependent and independent labels \cite{zhang2019lrfs}, DCR-MFS conditions relevance on selected features and on other labels \cite{zhang2021dcr}, and STFS unifies many low-order criteria \cite{gao2023stfs}. Recent work in this journal measures relevance through strongly relevant label gain and label mutual aid \cite{dai2024gain}, selects features by gradient-based optimisation that accounts for three-way variable interaction \cite{zou2024gradient}, and exploits the internal information of the candidate feature space \cite{liu2026candidate}. Neighbourhood-based measures handle continuous features: neighbourhood mutual information \cite{lin2016neighborhood}, adaptive fuzzy neighbourhood mutual information with label correlation \cite{sun2025afifc}, and a noise-tolerant neighbourhood discrimination index over transitive label pairs \cite{kang2026transitive}. These criteria differ in how individual terms are approximated, but aggregate relevance over labels by (weighted) summation, i.e.\ they are utilitarian in the sense of Section~\ref{sec:framework}.

\paragraph{Balancing and specialising relevance across labels}
Gonz\'alez-L\'opez et al.\ \cite{gonzalez2020} compare the Euclidean norm (ENM) and the geometric mean (GMM) of the per-label mutual informations. LSRIFS \cite{han2025lsrifs} multiplies relevance by an exponential of the squared mutual information to favour features that are informative about few labels. RRFS \cite{han2024rrfs} weights labels inversely to the information they have already received, and CIRFS \cite{liu2026cirfs} names the phenomenon \emph{coverage imbalance} and counters it with inverse-square-root weights interpolated with label frequencies. Proposition~\ref{prop:unify} locates these criteria on the $\alpha$-fair spectrum. Other works weight relevance by label importance \cite{ma2025label} or by class imbalance \cite{liu2025class}; in our framework, prior knowledge of label importance enters through the label weights $w_l$.

\paragraph{Embedded and multi-view methods}
Embedded methods learn a sparse linear map from features to labels and rank features by the row norms of the coefficient matrix. RFS \cite{nie2010rfs} combines an $\ell_{2,1}$-norm loss with an $\ell_{2,1}$-norm penalty, MIFS \cite{jian2016mifs} regresses onto a latent label space regularised by the feature-space manifold, and LLSF \cite{huang2016llsf} learns label-specific sparse coefficients coupled through label correlations. Recent embedded methods learn robust structures from discriminative label correlations \cite{jia2024discriminative} or select features via orthogonal regression \cite{xu2025orthogonal}. For data described by several views, MSFS \cite{zhang2020msfs} combines a latent label space with view-weighted manifold regularisation and view-wise sparsity; more recently, embedded feature fusion \cite{hao2025embedded} and graph-fusion-guided data reconstruction \cite{chen2026graph} have been proposed. The embedded methods compared here (RFS, MIFS and LLSF) minimise an unweighted aggregate of the fitting errors of all labels; how such an aggregate allocates the selected features among labels has not been analysed.

\paragraph{Submodularity and fairness}
Information gain is submodular under conditional independence \cite{krause2005}. When submodularity holds only approximately, greedy selection retains a constant-factor guarantee that depends on the submodularity ratio \cite{das2011}, also for objectives that are not submodular at all \cite{bian2017}, whereas maximising the worst of several submodular objectives is inapproximable \cite{krause2008robust}. ``Fair feature selection'' usually means avoiding dependence on a sensitive attribute \cite{khodadadian2021}; our notion concerns the allocation of information among labels and follows the $\alpha$-fair resource allocation of Mo and Walrand \cite{mo2000}. Welfare aggregation has recently been used to select one feature set for several sample populations \cite{lyu2026popfs} and to weight task losses in multi-task training \cite{ban2024fair}. To our knowledge, no multi-label feature selection method formulates the aggregation over labels as a welfare function of their information, analyses its submodularity, or relates it to per-label Bayes error.

\section{Multi-label feature selection as fair allocation of information}\label{sec:framework}

\subsection{Notation and background}\label{sec:prelim}

We observe $n$ instances with $d$ discrete (or discretised) features $X_1,\dots,X_d$ and $q$ binary labels $Y_1,\dots,Y_q$; $X_S=(X_j)_{j\in S}$ for $S\subseteq[d]$. Label $l$ has prior $p_l$, minority-class rate $\pi_l=\min\{p_l,1-p_l\}$ and entropy $H_l=H_b(p_l)$, where $H_b(t)=-t\log t-(1-t)\log(1-t)$ and $H_b^{-1}:[0,1]\to[0,\frac12]$ is its inverse on $[0,\frac12]$. Logarithms are base~2. We use the mutual information (MI) $I(X;Y)$ and the conditional mutual information (CMI) $I(X;Y\mid Z)$ \cite{brown2012}; by the chain rule $I(X_{S\cup\{j\}};Y)=I(X_S;Y)+I(X_j;Y\mid X_S)$, so $S\mapsto I(X_S;Y)$ is non-decreasing. A set function $F$ is submodular if $F(A\cup\{j\})-F(A)\ge F(B\cup\{j\})-F(B)$ for all $A\subseteq B$, $j\notin B$; for normalised, monotone, submodular $F$ greedy selection of $k$ elements attains $(1-1/e)\max_{|S|\le k}F(S)$ \cite{nemhauser1978}. The $\alpha$-fair utility \cite{mo2000} is $u_\alpha(x)=x^{1-\alpha}/(1-\alpha)$ for $\alpha\neq1$ and $u_1(x)=\log x$; maximising $\sum_lu_\alpha(x_l)$ over allocations gives the utilitarian solution for $\alpha=0$, proportional fairness for $\alpha=1$ \cite{kelly1998} and max--min fairness as $\alpha\to\infty$, while for $\alpha<0$ the utility is convex and rewards concentration.

\subsection{Labels as agents: the \texorpdfstring{$\alpha$}{alpha}-fair coverage objective}

A feature subset $S$ delivers to label $l$ the information $I(X_S;Y_l)\in[0,H_l]$. We call
\begin{equation}\label{eq:coverage}
a_l(S)=I(X_S;Y_l)\ \ \text{(raw)},\qquad
c_l(S)=I(X_S;Y_l)/H_l\in[0,1]\ \ \text{(normalised)}
\end{equation}
the coverage of label $l$ and write $\kappa_l(S)$ for either. The normalised coverage is Theil's uncertainty coefficient \cite{theil1970}, the fraction of the label's uncertainty resolved by $X_S$; raw coverage favours high-entropy (frequent) labels. The target is the \emph{vector} of per-label coverages rather than the joint information $I(X_S;Y_1,\dots,Y_q)$ approximated by PMU \cite{lee2013pmu}, which matches per-label evaluation metrics. Because the coverages compete for a common budget of $k$ features, every criterion implicitly decides \emph{which labels receive information}. To make this allocation explicit, we score a subset by the $\alpha$-fair welfare of the coverages. With a smoothing constant $\varepsilon>0$, let
\begin{equation}\label{eq:phi}
\phi_{\alpha}(x)=u_\alpha(x+\varepsilon)-u_\alpha(\varepsilon)=
\begin{cases}
\big[(x+\varepsilon)^{1-\alpha}-\varepsilon^{1-\alpha}\big]/(1-\alpha), & \alpha\neq1,\\
\log(1+x/\varepsilon), & \alpha=1,
\end{cases}
\end{equation}
and define the \emph{$\alpha$-fair label-information coverage}
\begin{equation}\label{eq:objective}
\Phi_\alpha(S)=\sum_{l=1}^{q} w_l\,\phi_\alpha\big(\kappa_l(S)\big),\qquad\text{maximise }\Phi_\alpha(S)\ \text{ s.t. }|S|\le k,
\end{equation}
with label weights $w_l\ge0$ (uniform unless prior knowledge says otherwise). $\Phi_\alpha$ is normalised and monotone, and $\phi'_\alpha(x)=(x+\varepsilon)^{-\alpha}$ is the \emph{marginal value of one bit} delivered to a label with coverage $x$: for $\alpha=0$ every bit is worth the same and $\Phi_0$ is the total coverage; for $\alpha>0$ a bit is worth less to a well-covered label, so information is steered toward under-served labels ($\alpha=1$ maximises the geometric mean of the smoothed coverages); as $\alpha\to\infty$ only the worst-covered label matters; and for $\alpha<0$ the selection keeps feeding labels that are already well served.

\subsection{A unifying view of existing criteria}\label{sec:unify}

Greedy information-theoretic filters add, at every step, the candidate $f$ with the largest score $J(f\mid S)$. Linearising the gain of Eq.~(\ref{eq:objective}) at the current coverage gives the \emph{first-order $\alpha$-fair score}
\begin{equation}\label{eq:firstorder}
J_\alpha(f\mid S)=\sum_{l=1}^q w_l\,\phi'_\alpha\big(\hat\kappa_l(S)\big)\,\hat g_l(f\mid S),
\end{equation}
where $\hat g_l(f\mid S)$ estimates the increment $\kappa_l(S\cup\{f\})-\kappa_l(S)$ and $\hat\kappa_l(S)$ is the coverage accumulated so far.

\begin{proposition}[Unification]\label{prop:unify}
Let $I_{fl}=I(X_f;Y_l)$ and use raw coverage with $w_l\equiv1$.
(i) For $\alpha=0$, $J_0(f\mid S)=\sum_l\hat g_l(f\mid S)$; with $\hat g_l=I_{fl}$ this is the relevance term of MIM, PMU \cite{lee2013pmu}, D2F \cite{lee2015d2f}, SCLS \cite{lee2017scls} and MDMR \cite{lin2015mdmr}, and with $\hat g_l=\frac1{|S|}\sum_{s\in S}I(X_f;Y_l\mid X_s)$ the feature-conditional relevance of STFS \cite{gao2023stfs} and DCR-MFS \cite{zhang2021dcr}, whose label-pair terms are likewise unweighted sums over labels.
(ii) For $\alpha=1$, $\varepsilon\to0$, $\hat g_l=I_{fl}$ and the ledger $\hat\kappa_l(S)=A_l(S)=\sum_{s\in S}I_{sl}$, $J_1(f\mid S)=\sum_l I_{fl}/A_l(S)=T(S)^{-1}\sum_l \mathrm{LC}_l\,I_{fl}$, where $\mathrm{LC}_l=T(S)/A_l(S)$ and $T(S)=\sum_l A_l(S)$ are the label information coefficients of RRFS \cite{han2024rrfs}; the factor $T(S)^{-1}$ is common to all candidates. The weights $(A_l(S)+\varepsilon)^{-1/2}=\phi'_{1/2}(A_l(S))$ of CIRFS \cite{liu2026cirfs} are the first-order weights of $\alpha=1/2$, which CIRFS interpolates with label frequencies.
(iii) For $\alpha=1$ and $S=\emptyset$, $\Phi_1(\{f\})=\sum_l\log(1+I_{fl}/\varepsilon)$; if all $I_{fl},I_{f'l}>0$ and $\prod_lI_{fl}\ne\prod_lI_{f'l}$, then for all sufficiently small $\varepsilon$ the order of $f,f'$ is their order under the geometric mean of GMM \cite{gonzalez2020}.
(iv) For $\alpha=-1$ and $S=\emptyset$, $\Phi_{-1}(\{f\})=\frac12\sum_lI_{fl}^2+\varepsilon\sum_lI_{fl}$, whose ranking tends to that of the Euclidean norm of ENM \cite{gonzalez2020} as $\varepsilon\to0$; the LSRIFS relevance $\theta_f\cdot\frac1q\sum_lI_{fl}$ with $\theta_f=\exp(\sum_lI_{fl}^2/Z)$ \cite{han2025lsrifs} is the product of a monotone transform of the $\alpha=-1$ score and the $\alpha=0$ score.
\end{proposition}

\begin{proof}
(i), (ii) Substitute $\phi'_0\equiv1$, $\phi'_1(x)=1/(x+\varepsilon)$ and $\phi'_{1/2}(x)=(x+\varepsilon)^{-1/2}$ into Eq.~(\ref{eq:firstorder}). (iii) $\Phi_1(\{f\})-\Phi_1(\{f'\})\to\log\prod_lI_{fl}-\log\prod_lI_{f'l}\neq0$ as $\varepsilon\to0$. (iv) Expand $\phi_{-1}(x)=\frac12[(x+\varepsilon)^2-\varepsilon^2]$.
\end{proof}

\begin{figure}[!t]
\centering
\includegraphics[width=\linewidth]{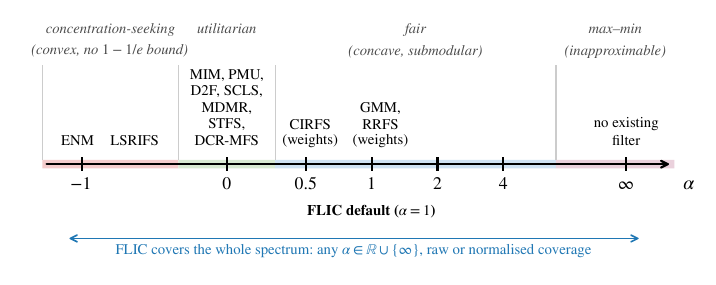}
\caption{The $\alpha$-fair spectrum of multi-label relevance criteria (Proposition~\ref{prop:unify}). Concentration-seeking criteria ($\alpha<0$) need not be submodular, and the $(1-1/e)$ guarantee fails for them; every $\alpha\ge0$ yields a submodular objective with a $(1-1/e)$ greedy guarantee whenever each label's information is submodular (Theorem~\ref{thm:submod}); the max--min limit is inapproximable (Proposition~\ref{prop:maxmin}).}
\label{fig:spectrum}
\end{figure}

Proposition~\ref{prop:unify}, Table~\ref{tab:taxonomy} and Fig.~\ref{fig:spectrum} place the literature on one axis: the opposite recommendations---to favour features whose information is concentrated on a few labels (ENM \cite{gonzalez2020}, LSRIFS \cite{han2025lsrifs}) and to balance the information among labels (GMM \cite{gonzalez2020}, RRFS \cite{han2024rrfs}, CIRFS \cite{liu2026cirfs})---are different values of $\alpha$. The view also exposes two gaps: no filter targets the max--min end of the spectrum, and none combines the \emph{exact} gain of a concave welfare with \emph{label-wise} conditional information.

\begin{table}[!t]
\centering
\caption{Multi-label relevance criteria as points on the $\alpha$-fair spectrum (Proposition~\ref{prop:unify}). All existing criteria use raw coverage. ``Gain'' is the per-label increment estimate $\hat g_l(f\mid S)$, with $I=I(X_f;Y_l)$; ``Ledger'' is the coverage $\hat\kappa_l(S)$ that drives the label weights.}
\label{tab:taxonomy}
\footnotesize\linespread{1}\selectfont
\setlength{\tabcolsep}{3pt}
\renewcommand{\arraystretch}{1.25}
\begin{adjustbox}{max width=\linewidth}\begin{tabular}{@{}>{\raggedright\arraybackslash}p{2.35cm}>{\raggedright\arraybackslash}p{2.2cm}l>{\raggedright\arraybackslash}p{2.45cm}@{}}
\toprule
Criterion & Implied $\alpha$ & Gain $\hat g_l(f\mid S)$ & Ledger / remarks\\
\midrule
MIM, PMU, D2F, SCLS, MDMR & $0$ & $I(X_f;Y_l)$ & none\\
STFS, DCR-MFS & $0$ & $\overline{I}(X_f;Y_l\mid X_s)$ & $+$ label-pair terms\\
ENM & $-1$ & $I(X_f;Y_l)$ & static ranking\\
LSRIFS & $-1$ and $0$ (product) & $I(X_f;Y_l)$ & static weight $\theta_f$\\
GMM & $1$ & $I(X_f;Y_l)$ & static ranking\\
RRFS & $1$ (first order) & $I(X_f;Y_l)$ & $\sum_{s\in S}I(X_s;Y_l)$\\
CIRFS & $1/2$ (first order) $+$ label frequency & $I+\gamma\,\overline{I}(X_f;Y_l\mid Y_j)/\sqrt{H_f}$ & $\sum_{s\in S}I(X_s;Y_l)$\\
\midrule
\textbf{FLIC} (raw or normalised) & any $\alpha\in\mathbb{R}\cup\{\infty\}$ & $\min\{I,\min_{s\in S}I(X_f;Y_l\mid X_s)\}$ & chain-rule ledger, exact gain\\
\bottomrule
\end{tabular}\end{adjustbox}
\end{table}

\subsection{Theoretical properties}\label{sec:theory}

\begin{theorem}[Greedy guarantee]\label{thm:submod}
Suppose every $F_l(S)=I(X_S;Y_l)$ is submodular, as holds when the features are conditionally independent given $Y_l$ \cite{krause2005}. Then for every $\alpha\ge0$, $\varepsilon>0$, $w\ge0$ and either coverage, $\Phi_\alpha$ is normalised, monotone and submodular, and greedy selection with exact marginal gains returns $S_k$ with $\Phi_\alpha(S_k)\ge(1-1/e)\max_{|S|\le k}\Phi_\alpha(S)$. For $\alpha<0$ the conclusion fails in general: for small $\varepsilon$, greedy selection can attain only about a fraction $k^{\alpha}$ of the optimum.
\end{theorem}
\begin{proof}
$\kappa_l$ is $F_l$ up to a positive factor, and $\phi_\alpha$ is non-decreasing and, for $\alpha\ge0$, concave ($\phi''_\alpha(x)=-\alpha(x+\varepsilon)^{-\alpha-1}\le0$). For $A\subseteq B$, $j\notin B$, $a=\kappa_l(A)\le b=\kappa_l(B)$ and increments $\delta_A\ge\delta_B\ge0$ of $\kappa_l$ at $A$ and $B$, monotonicity and concavity give $\phi_\alpha(a+\delta_A)-\phi_\alpha(a)\ge\phi_\alpha(a+\delta_B)-\phi_\alpha(a)\ge\phi_\alpha(b+\delta_B)-\phi_\alpha(b)$, so every term of $\Phi_\alpha$ is monotone submodular, and so is their sum; the bound is that of \cite{nemhauser1978}. For $\alpha<0$, take one label and two features with $F(\{1\})=F(\{2\})=x$ and $F(\{1,2\})=2x-r$, $0\le r\le x$ (a submodular $F$). With $\alpha=-1$ and $\varepsilon\to0$ the gain of feature 2 is $x^2/2$ at $\emptyset$ but $\frac12[(2x-r)^2-x^2]$ at $\{1\}$, which is larger whenever $r<(2-\sqrt2)x$---for instance for two conditionally independent weak proxies of the same label. Supplementary Section~S1 gives instances of the last statement in which every $F_l$ is submodular.
\end{proof}

When submodularity holds only approximately, the guarantee degrades gracefully, but fairness has a price: if every $F_l$ has submodularity ratio at least $\gamma$ \cite{das2011}, the ratio of $\Phi_\alpha$ is at least $\gamma(\varepsilon/(\bar\kappa+\varepsilon))^{|\alpha|}$, where $\bar\kappa$ bounds the coverages ($\max_lH_l$ for raw and $1$ for normalised coverage), so greedy retains a $1-\exp[-\gamma(\varepsilon/(\bar\kappa+\varepsilon))^{|\alpha|}]$ fraction of the optimum by the results of Das and Kempe \cite{das2011} and Bian et al.\ \cite{bian2017} (Supplementary Proposition~S1). For $\alpha<0$ this bound, with $\gamma=1$, still holds, but it vanishes as $\varepsilon\to0$. The fully fair limit is intractable; proofs of the next two results are given in Supplementary Section~S1.

\begin{proposition}[Max--min limit]\label{prop:maxmin}
For a finite family $\mathcal{S}$ of feasible subsets and $w_l\equiv1$, there is $\alpha_0$ such that for all $\alpha>\alpha_0$ every maximiser of $\Phi_\alpha$ over $\mathcal{S}$ is leximin-optimal for $(\kappa_l(S)+\varepsilon)_l$, and in particular maximises $\min_l\kappa_l(S)$. Conversely, $\max_{|S|\le k}\min_lF_l(S)$ with monotone submodular $F_l$ admits no polynomial-time approximation within any positive factor unless P$\,=\,$NP \cite{krause2008robust}, even when every $F_l$ is an information function $I(X_S;Y_l)$.
\end{proposition}

Why should one care about coverage at all? The next result ties both scales to the error of the best possible predictor of each label.

\begin{theorem}[Per-label error bounds]\label{thm:bounds}
For every label $l$ and subset $S$, the Bayes error $P^\star_l(S)$ of predicting $Y_l$ from $X_S$ satisfies
\begin{equation}\label{eq:fano-hr}
H_b^{-1}\!\big(H_l\,(1-c_l(S))\big)\ \le\ P^\star_l(S)\ \le\ \tfrac12\,H_l\,(1-c_l(S))=\tfrac12\big(H_l-a_l(S)\big).
\end{equation}
Consequently, (a) the Bayes Hamming loss satisfies $H_b^{-1}(R(S)/q)\le\mathrm{HL}^\star(S)\le R(S)/(2q)$ with $R(S)=\sum_lH_l-\sum_la_l(S)$, so both bounds improve monotonically with the utilitarian raw objective $\sum_la_l(S)$; and (b) the relative error $r_l(S)=P^\star_l(S)/\pi_l$, the fraction of the error of the trivial majority predictor that survives, satisfies
\begin{equation}\label{eq:psi}
r_l(S)\ \ge\ \psi_{\pi_l}\big(1-c_l(S)\big),\qquad \psi_\pi(t)=H_b^{-1}\big(t\,H_b(\pi)\big)/\pi,
\end{equation}
where $\psi_\pi$ is increasing and convex with $\psi_\pi(0)=0$, $\psi_\pi(1)=1$ and $\psi_\pi(t)\to t$ as $\pi\to0$.
\end{theorem}

The bounds in Eq.~(\ref{eq:fano-hr}) are Fano's \cite{cover2006} and the Hellman--Raviv \cite{hellman1970} inequalities applied to $H(Y_l\mid X_S)=H_l(1-c_l(S))$. Raw coverage is thus the currency of the \emph{Hamming loss}, whose Bayes value is sandwiched between increasing functions of the total residual entropy, which $\alpha=0$ with raw coverage minimises. Normalised coverage is the currency of \emph{label-wise} quality: a rare label $l$ with normalised coverage $c$ keeps, whatever the classifier, at least a $\psi_{\pi_l}(1-c)$ fraction of its trivial error, which tends to $1-c$ as $\pi_l\to0$ (Supplementary Fig.~S1). Because $\psi_\pi$ is convex, an increment of coverage lowers this bound the more, the less covered the label already is, which favours steering information toward poorly covered labels---the purpose of $\alpha>0$. The theory therefore suggests that small $\alpha$ favours Hamming loss and moderate $\alpha$ label-wise metrics.

\section{The FLIC algorithm}\label{sec:algorithm}

Maximising Eq.~(\ref{eq:objective}) requires the coverage $\kappa_l(S\cup\{f\})$ of every label for every candidate $f$, i.e.\ the high-order quantity $I(X_{S\cup\{f\}};Y_l)$, which cannot be estimated reliably from a few thousand instances once $|S|$ exceeds a handful of discretised features. FLIC follows the information-theoretic tradition of replacing high-order terms by low-order surrogates \cite{brown2012}, but does so \emph{label by label} and inside an exact accounting of the welfare gain.

\paragraph{Chain-rule coverage ledger}
By the chain rule, adding $f$ to $S$ increases the information about label $l$ by exactly $I(X_f;Y_l\mid X_S)$. FLIC therefore keeps for every label a ledger of the information accumulated along the greedy path,
\begin{equation}\label{eq:ledger}
\hat a_l(S\cup\{f\})=\min\big\{\hat a_l(S)+\hat g_l(f\mid S),\ H_l\big\},
\end{equation}
where $\hat g_l(f\mid S)$ estimates $I(X_f;Y_l\mid X_S)$ and the cap enforces $a_l\le H_l$. With $S=\{s_1,\dots,s_t\}$ we use the label-wise conditional-minimum estimate
\begin{equation}\label{eq:cmim}
\hat g_l(f\mid S)=\min\Big\{\tilde I(X_f;Y_l),\ \min_{s\in S}\tilde I(X_f;Y_l\mid X_s)\Big\}\qquad(\hat g_l(f\mid\emptyset)=\tilde I(X_f;Y_l)),
\end{equation}
in the spirit of CMIM \cite{fleuret2004} but applied separately to each label. Conditioning on a single selected feature captures the most redundant pair; the minimum upper-bounds $I(X_f;Y_l\mid X_S)$ when conditioning on more variables creates no synergy, and it never increases as $S$ grows.

\paragraph{Bias correction}
Plug-in estimates of mutual information are biased upwards by about $(|\mathcal{X}|-1)(|\mathcal{Y}|-1)/(2n\ln2)$ bits, as derived by Miller \cite{miller1955} and analysed for finite samples by Panzeri and Treves \cite{panzeri1996}. Summed along the greedy path, this bias would make every ledger grow even when only noise is selected. FLIC therefore uses $\tilde I(X_f;Y_l)=\max\{0,\hat I(X_f;Y_l)-b_f\}$ and $\tilde I(X_f;Y_l\mid X_s)=\max\{0,\hat I(X_f;Y_l\mid X_s)-|\mathcal{X}_s|\,b_f\}$ with $b_f=(|\mathcal{X}_f|-1)/(2n\ln2)$.

\paragraph{Exact welfare gains}
Given the ledger, the score of candidate $f$ is the exact increase of the welfare under the estimated increments,
\begin{equation}\label{eq:score}
\Delta(f\mid S)=\sum_{l=1}^qw_l\Big[\phi_\alpha\big(\min\{\hat\kappa_l(S)+\hat g_l(f\mid S)/\nu_l,\ \bar\kappa_l\}\big)-\phi_\alpha\big(\hat\kappa_l(S)\big)\Big],
\end{equation}
with $\nu_l=1$, $\bar\kappa_l=H_l$ for raw and $\nu_l=H_l$, $\bar\kappa_l=1$ for normalised coverage. Unlike the first-order score Eq.~(\ref{eq:firstorder}), the exact gain accounts for saturation: a feature that would push a label beyond its entropy, or adds much to an already well-covered label, is valued correctly by the concave utility. For $\alpha=\infty$ FLIC compares the sorted vectors of updated coverages lexicographically (leximin). Algorithm~\ref{alg:flic} summarises the procedure.

\begin{algorithm}[!t]
\caption{FLIC: Fair Label-Information Coverage}\label{alg:flic}
\begin{algorithmic}[1]
\Require discretised training data $X\in\mathbb{N}^{n\times d}$, labels $Y\in\{0,1\}^{n\times q}$, budget $k$, fairness $\alpha$, smoothing $\varepsilon$, label weights $w$
\Ensure ranked feature subset $S=(s_1,\dots,s_k)$
\State $H_l$ for all $l$; \ $G_{fl}\gets\tilde I(X_f;Y_l)$ for all $f,l$ \Comment{one matrix product}
\State $\hat\kappa_l\gets0$ for all $l$; \ $S\gets\emptyset$
\For{$t=1,\dots,k$}
  \State $\Delta_f\gets\sum_lw_l\big[\phi_\alpha(\min\{\hat\kappa_l+G_{fl}/\nu_l,\bar\kappa_l\})-\phi_\alpha(\hat\kappa_l)\big]$ for $f\notin S$
  \State $s_t\gets\arg\max_{f\notin S}\Delta_f$; \ $S\gets S\cup\{s_t\}$
  \State $\hat\kappa_l\gets\min\{\hat\kappa_l+G_{s_tl}/\nu_l,\bar\kappa_l\}$ for all $l$ \Comment{ledger, Eq.~(\ref{eq:ledger})}
  \State $C_{fl}\gets\tilde I(X_f;Y_l\mid X_{s_t})$ for all $f,l$ \Comment{one matrix product}
  \State $G_{fl}\gets\min\{G_{fl},C_{fl}\}$ for all $f,l$ \Comment{label-wise CMIM, Eq.~(\ref{eq:cmim})}
\EndFor
\State \Return $S$
\end{algorithmic}
\end{algorithm}

\begin{proposition}\label{prop:flic}
(i) \emph{Deduplication without a trade-off parameter.} If a candidate $f$ is a deterministic function of a selected feature $s$ (e.g.\ a duplicate), then $\hat I(X_f;Y_l\mid X_s)=0$ for every $l$, so $\Delta(f\mid S)=0$.
(ii) \emph{Diminishing surrogate returns.} For $\alpha\ge0$ the score $\Delta(f\mid S)$ of every candidate is non-increasing along the greedy path, mirroring Theorem~\ref{thm:submod}.
(iii) \emph{Complexity.} With $B$ discretisation bins, each iteration costs one $(dB\times n)\cdot(n\times 2qB)$ matrix product plus $O(dqB^2)$ work, i.e.\ $O(kndq)$ time and $O(dq+ndB)$ memory overall.
\end{proposition}
\begin{proof}
(i) If $X_f=h(X_s)$ then $H(X_f\mid X_s)=0$ and $I(X_f;Y_l\mid X_s)=0$; this holds exactly for the plug-in estimate, and the bias correction keeps it at zero. (ii) $\hat g_l(f\mid S)$ is a minimum over a growing set and $\hat\kappa_l(S)$ does not decrease; for concave non-decreasing $\phi_\alpha$ the increment $\phi_\alpha(\min\{x+\delta,\bar\kappa\})-\phi_\alpha(x)$ is non-decreasing in $\delta$ and non-increasing in $x$. (iii) With $X$ one-hot encoded ($n\times dB$) and the pairs $(Y_l,X_{s_t})$ one-hot encoded ($n\times 2qB$), all three-way contingency tables are entries of one matrix product.
\end{proof}

Property (i) contrasts with criteria that subtract $\beta\sum_sI(X_f;X_s)$, where a duplicate of a highly relevant feature can still win and the appropriate $\beta$ changes with $|S|$ and $q$ \cite{zhang2021dcr}. In FLIC redundancy is measured in bits of information about each label, so no balancing parameter is needed, and a feature that is redundant for one label but new for another is credited only for the latter. The only parameters are $\alpha$, the desired fairness, and $\varepsilon$, the coverage level (in bits) below which differences between labels are not resolved.

\section{Experimental setup}\label{sec:exp}

The experiments address four questions. (Q1) Does FLIC select better feature subsets than state-of-the-art information-theoretic filters, and how does it compare with embedded methods? (Q2) Does $\alpha$ trade Hamming loss against label-wise quality as suggested by Theorem~\ref{thm:bounds}? (Q3) Which components of FLIC matter? (Q4) Are the conclusions robust to the classifier and to multi-view data?

\paragraph{Data sets}
We use 22 benchmark data sets: eleven from the Mulan repository \cite{tsoumakas2011mulan}, covering music (emotions, CAL500), images (scene, flags), biology (yeast, genbase), audio (birds) and text (medical, enron, bibtex, tmc2007-500), and the eleven Yahoo! web-page collections of Ueda and Saito \cite{ueda2003pmm} in the reduced-vocabulary version of Zhang and Zhou \cite{zhang2007mlknn}, whose training and test files are pooled. Constant features and labels with fewer than five positive instances are removed (Table~\ref{tab:datasets}); the label priors span three orders of magnitude. To avoid tuning on test data, all design choices of FLIC (coverage scale, $\alpha$, $\varepsilon$) and the regularisation parameters of the embedded baselines were fixed on four \emph{separate} development data sets---Corel5k, Stackex-chemistry, Ohsumed and EukaryotePseAAC, each subsampled to 3000 instances---which are not used in any reported comparison (Supplementary Section~S2).

\begin{table}[!t]
\centering
\caption{Characteristics of the 22 evaluation data sets after removing constant features and labels with fewer than five positive instances. Card.: label cardinality; Dens.: label density; $\pi_{\min}$: smallest label prior; $k_{\max}$: largest subset size evaluated.}
\label{tab:datasets}
\footnotesize
\setlength{\tabcolsep}{4pt}
\begin{adjustbox}{max width=\linewidth}\begin{tabular}{llrrrrrrr}
\toprule
Data set & Domain & $n$ & $d$ & $q$ & Card. & Dens. & $\pi_{\min}$ & $k_{\max}$\\
\midrule
emotions & Music & 593 & 72 & 6 & 1.87 & 0.311 & 0.2496 & 15\\
flags & Image & 194 & 19 & 7 & 3.39 & 0.485 & 0.1340 & 10\\
scene & Image & 2407 & 294 & 6 & 1.07 & 0.179 & 0.1512 & 59\\
yeast & Biology & 2417 & 103 & 14 & 4.24 & 0.303 & 0.0141 & 21\\
birds & Audio & 645 & 260 & 19 & 1.01 & 0.053 & 0.0093 & 52\\
genbase & Biology & 662 & 111 & 19 & 1.23 & 0.065 & 0.0076 & 23\\
medical & Text & 978 & 1449 & 25 & 1.21 & 0.048 & 0.0061 & 290\\
enron & Text & 1702 & 1001 & 49 & 3.37 & 0.069 & 0.0035 & 201\\
CAL500 & Music & 502 & 68 & 174 & 26.04 & 0.150 & 0.0100 & 14\\
bibtex & Text & 7395 & 1836 & 159 & 2.40 & 0.015 & 0.0069 & 368\\
tmc2007-500 & Text & 28596 & 500 & 22 & 2.22 & 0.101 & 0.0141 & 100\\
Arts & Text & 5000 & 462 & 25 & 1.64 & 0.065 & 0.0010 & 93\\
Business & Text & 5000 & 438 & 27 & 1.59 & 0.059 & 0.0014 & 88\\
Computers & Text & 5000 & 681 & 30 & 1.51 & 0.050 & 0.0010 & 137\\
Education & Text & 5000 & 550 & 27 & 1.46 & 0.054 & 0.0016 & 110\\
Entertainment & Text & 5000 & 640 & 20 & 1.42 & 0.071 & 0.0010 & 128\\
Health & Text & 5000 & 612 & 24 & 1.66 & 0.069 & 0.0012 & 123\\
Recreation & Text & 5000 & 606 & 22 & 1.42 & 0.065 & 0.0038 & 122\\
Reference & Text & 5000 & 793 & 24 & 1.16 & 0.049 & 0.0010 & 159\\
Science & Text & 5000 & 743 & 38 & 1.45 & 0.038 & 0.0012 & 149\\
Social & Text & 5000 & 1047 & 35 & 1.28 & 0.037 & 0.0012 & 210\\
Society & Text & 5000 & 636 & 26 & 1.69 & 0.065 & 0.0014 & 128\\
\bottomrule
\end{tabular}\end{adjustbox}
\end{table}

\paragraph{Compared methods}
FLIC is compared with thirteen information-theoretic filters spanning the spectrum of Table~\ref{tab:taxonomy}---MIM (summed mutual information) \cite{brown2012}, PMU \cite{lee2013pmu}, D2F \cite{lee2015d2f}, SCLS \cite{lee2017scls}, MDMR \cite{lin2015mdmr}, LRFS \cite{zhang2019lrfs}, DCR-MFS (the averaged variant DCR-MFSarg) \cite{zhang2021dcr}, STFS \cite{gao2023stfs}, GMM and ENM \cite{gonzalez2020}, LSRIFS \cite{han2025lsrifs}, RRFS \cite{han2024rrfs} and CIRFS \cite{liu2026cirfs} (with the parameter values recommended by its authors)---and with three embedded methods, RFS \cite{nie2010rfs}, MIFS \cite{jian2016mifs} and LLSF \cite{huang2016llsf}. All filters were re-implemented from the formulas of the original papers on top of one vectorised, unit-tested estimator of (conditional) mutual information, so that they share the same discretisation and probability estimates (Supplementary Section~S3); their comparison therefore isolates the selection criterion. The one exception is the bias correction of Section~\ref{sec:algorithm}, which is part of FLIC; the baselines use the plug-in estimates of their original formulations, and a control in which they use the same correction is reported in Section~\ref{sec:main}. The author developed the implementation and the experiment and analysis scripts together with the AI coding assistant Claude Code (Anthropic), running the Claude Opus~5.5 model, and reviewed all code and results in full; unit tests in the code check the information estimator and the ML-kNN classifier against direct brute-force implementations, the greedy selections of every filter against a brute-force implementation of its formula in Supplementary Section~S3, and the embedded baselines on planted features. The embedded methods, which see the continuous feature values, are compared separately; they are fitted on standardised continuous features and rank features by the row norms of the coefficient matrix; their regularisation parameters were selected on the development data from grids of 5 (RFS), 49 (MIFS) and 15 (LLSF) configurations (Supplementary Section~S3). FLIC uses the setting fixed on the development data: raw coverage, $\alpha=1$, $\varepsilon=0.2$ bit, uniform label weights, the label-wise conditional-minimum gain of Eq.~(\ref{eq:cmim}) and bias correction; variants on the normalised scale, which is dimensionless, use $\varepsilon=0.1$. On the development data, raw-scale settings with $\alpha\le1$ and $\varepsilon\ge0.1$ were nearly tied (average ranks 15.1--16.6, $\alpha=0$ best); we chose the canonical proportional fairness $\alpha=1$, which ranked best on Tail-AUC (Supplementary Section~S2).

\paragraph{Protocol and metrics}
Every data set is split into five folds by iterative stratification \cite{sechidis2011}. On each training fold, continuous features are discretised into three equal-width bins (features with at most three distinct values are kept), as in the evaluations of DCR-MFS \cite{zhang2021dcr} and RRFS \cite{han2024rrfs}, and every method ranks $k_{\max}=\max\{\lceil0.2d\rceil,\min(d,10)\}$ features. For 20 equally spaced subset sizes $k\le k_{\max}$ (fewer if $k_{\max}<20$), an ML-kNN classifier \cite{zhang2007mlknn} ($10$ neighbours, smoothing $1$, Euclidean distance on min--max scaled features) is trained on the top-$k$ features and evaluated on the test fold. We report eight standard metrics \cite{zhang2014review}---Hamming loss (HL), ranking loss (RL), coverage (CV, divided by $q$), average precision (AP), one-error (OE), micro-F1 (MiF1), macro-F1 (MaF1) and macro-averaged AUC (MaAUC)---and \emph{Tail-AUC}, which we define as the mean AUC of the 10\% worst-predicted labels of each test fold. HL and MiF1 pool all instance--label pairs and are dominated by frequent labels, whereas MaF1 and MaAUC weight all labels equally and Tail-AUC considers only the worst-predicted ones; we call the latter three label-wise metrics. Each value is averaged over the subset sizes (the normalised area under the performance curve) and the five folds. Methods are compared by average ranks, the Friedman test with the Iman--Davenport correction, the Bonferroni--Dunn test with FLIC as control and two-sided Wilcoxon signed-rank tests (exact $p$-values) with Holm's correction \cite{demsar2006}; Table~\ref{tab:ranksfilters} gives the resulting ranks of the filters.

\section{Results}\label{sec:results}

\subsection{Comparison with information-theoretic filters (Q1)}\label{sec:main}

\begin{table}[!t]
\centering
\caption{Average ranks (lower is better) over the 22 data sets when the comparison is restricted to the 14 information-theoretic filters. ``Sig.'' gives the number of metrics on which FLIC is significantly better/worse than the filter (Wilcoxon signed-rank test, Holm correction over the 13 filters, $p<0.05$). Best rank in bold, second best underlined.}
\label{tab:ranksfilters}
\footnotesize
\setlength{\tabcolsep}{3pt}
\begin{adjustbox}{max width=\linewidth}\begin{tabular}{lccccccccccc}
\toprule
Method & HL & RL & CV & AP & OE & MiF1 & MaF1 & MaAUC & TailAUC & Mean & Sig.\\
\midrule
MIM & 8.11 & 8.45 & 8.50 & 8.68 & 8.95 & 8.18 & 8.09 & 8.86 & 8.55 & 8.49 & 8/0\\
PMU & 12.77 & 11.95 & 12.14 & 11.91 & 11.32 & 12.64 & 12.91 & 12.45 & 10.05 & 12.02 & 8/0\\
D2F & 11.18 & 10.64 & 10.59 & 10.45 & 10.05 & 10.95 & 11.18 & 10.73 & 9.41 & 10.58 & 8/0\\
SCLS & 7.05 & 7.14 & 6.95 & 6.95 & 7.41 & 7.27 & 7.41 & 6.73 & 7.36 & 7.14 & 7/0\\
MDMR & 6.73 & 6.50 & 6.41 & 6.91 & 7.05 & 7.36 & 6.41 & 6.45 & 7.32 & 6.79 & 8/0\\
LRFS & 7.23 & 7.09 & 6.91 & 7.27 & 7.41 & 6.86 & 7.00 & 6.82 & 6.86 & 7.05 & 8/0\\
DCR-MFS & \textbf{3.18} & \underline{4.18} & \underline{4.50} & \underline{3.59} & \underline{3.50} & \underline{4.14} & \underline{3.77} & \underline{4.55} & 5.36 & \underline{4.09} & 3/0\\
STFS & 8.89 & 9.14 & 9.09 & 9.36 & 10.00 & 8.73 & 9.23 & 9.05 & 8.32 & 9.09 & 8/0\\
GMM & 11.91 & 11.50 & 11.82 & 11.68 & 12.09 & 11.91 & 12.09 & 11.50 & 10.27 & 11.64 & 8/0\\
ENM & 5.64 & 6.41 & 6.50 & 6.32 & 6.32 & 5.86 & 5.55 & 6.91 & 8.59 & 6.45 & 7/0\\
LSRIFS & 4.68 & 4.64 & 4.55 & 4.59 & 4.32 & 4.59 & 4.05 & 5.09 & 6.95 & 4.83 & 0/0\\
RRFS & 8.86 & 8.36 & 7.55 & 8.82 & 8.95 & 8.91 & 7.73 & 6.45 & \textbf{3.82} & 7.72 & 8/0\\
CIRFS & \underline{4.27} & 6.27 & 6.64 & 5.41 & 4.41 & 4.41 & 6.91 & 6.73 & 7.14 & 5.80 & 5/0\\
\midrule
FLIC & 4.50 & \textbf{2.73} & \textbf{2.86} & \textbf{3.05} & \textbf{3.23} & \textbf{3.18} & \textbf{2.68} & \textbf{2.68} & \underline{5.00} & \textbf{3.32} & --\\
\bottomrule
\end{tabular}\end{adjustbox}
\end{table}

\begin{figure}[!t]
\centering
\begin{subfigure}{0.49\linewidth}\includegraphics[width=\linewidth]{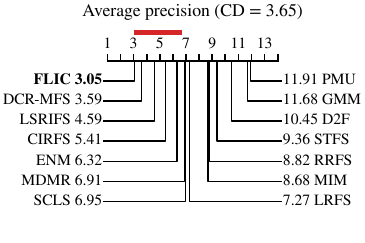}\end{subfigure}
\begin{subfigure}{0.49\linewidth}\includegraphics[width=\linewidth]{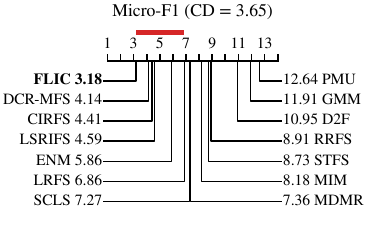}\end{subfigure}
\begin{subfigure}{0.49\linewidth}\includegraphics[width=\linewidth]{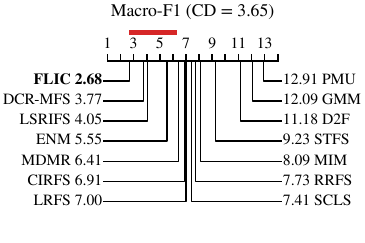}\end{subfigure}
\begin{subfigure}{0.49\linewidth}\includegraphics[width=\linewidth]{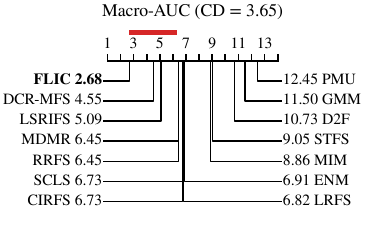}\end{subfigure}
\caption{Bonferroni--Dunn tests among the 14 information-theoretic filters with FLIC as the control (significance level 0.05, critical difference 3.65), for four of the seven metrics on which FLIC ranks first (the others follow from Table~\ref{tab:ranksfilters}). Filters whose average rank lies beyond the red bar (FLIC's rank plus the critical difference) are significantly worse than FLIC. Diagrams of all nine metrics for all 17 methods (critical difference 4.50) are given in Supplementary Fig.~S2.}
\label{fig:cdit}
\end{figure}

Table~\ref{tab:ranksfilters} reports the average ranks of FLIC and the thirteen information-theoretic filters; only FLIC corrects the estimation bias, a difference examined at the end of this section. The Iman--Davenport test rejects equal performance on every metric ($F_F$ between 5.33 and 24.26, critical value 1.76). FLIC is first on seven of the nine metrics---ranking loss, coverage, average precision, one-error, micro-F1, macro-F1 and macro-AUC---second on Tail-AUC, behind RRFS, and third on Hamming loss, behind DCR-MFS and CIRFS. Its mean rank over the nine metrics is 3.32, against 4.09 for the runner-up DCR-MFS and 4.83 for LSRIFS. Of the 117 pairwise Wilcoxon--Holm comparisons, 86 are significant in favour of FLIC and none against it: FLIC is significantly better than eleven of the thirteen filters on five to eight metrics each, including the recent coverage-aware CIRFS, and than DCR-MFS on ranking loss, coverage and macro-AUC, while no difference with LSRIFS is significant. The Bonferroni--Dunn test (Fig.~\ref{fig:cdit}) confirms the margin: on each of the seven metrics on which FLIC ranks first, it is significantly better than nine to eleven of the thirteen filters, and no filter is significantly better than FLIC on any metric. Nor is the lead an artefact of FLIC alone correcting the estimation bias: when all 13 filters use the same correction, FLIC keeps the best mean rank (3.20, against 3.94 for DCR-MFS), is first or joint first on six metrics and second on the other three, and 89 of the 117 comparisons are significant in its favour and none against it, with every corrected filter significantly worse than FLIC on at least three metrics (Supplementary Section~S9).

\subsection{Comparison with embedded methods}\label{sec:embedded}

\begin{table}[!t]
\centering
\caption{FLIC versus the embedded methods over the 22 data sets. Metric columns: number of data sets on which FLIC is better/worse (ML-kNN); $^{*}$ and $^{\dagger}$ mark a significant difference in favour of FLIC and of the embedded method (Wilcoxon signed-rank test, Holm correction over all 16 baselines, $p<0.05$). Last two columns: mean rank over the nine metrics among all 17 methods with ML-kNN and with logistic regression (lower is better; full tables in the Supplementary Material).}
\label{tab:embedded}
\footnotesize
\setlength{\tabcolsep}{2.5pt}
\begin{adjustbox}{max width=\linewidth}\begin{tabular}{lccccccccccc}
\toprule
Method & HL & RL & CV & AP & OE & MiF1 & MaF1 & MaAUC & TailAUC & kNN & LR\\
\midrule
RFS & 5/17 & 16/6 & 17/5 & 6/16 & 5/17$^{\dagger}$ & 5/17$^{\dagger}$ & 11/11 & 17/5$^{*}$ & 15/7 & 5.83 & 6.99\\
MIFS & 4/18 & 13/9 & 17/5 & 6/16 & 5/17 & 6/16 & 14/8 & 16/6$^{*}$ & 19/3$^{*}$ & 6.25 & 7.01\\
LLSF & 4/18 & 10/12 & 12/10 & 5/17 & 5/17$^{\dagger}$ & 4/18$^{\dagger}$ & 8/14 & 12/10 & 16/6 & 5.14 & 5.79\\
\midrule
FLIC & -- & -- & -- & -- & -- & -- & -- & -- & -- & \textbf{4.94} & \textbf{4.64}\\
\bottomrule
\end{tabular}\end{adjustbox}
\end{table}

Table~\ref{tab:embedded} compares FLIC with the three embedded methods, which select features from the continuous values rather than from three-bin discretisations. Over the nine metrics, FLIC has the best mean rank among all 17 methods with both classifiers---4.94 with ML-kNN and 4.64 with logistic regression, against 5.14 and 5.79 for the strongest embedded method, LLSF (Supplementary Tables~S12 and~S14). With ML-kNN, as with logistic regression, FLIC also has the best rank of all 17 methods on ranking loss (3.95), coverage (3.77) and macro-AUC (3.64), although against LLSF alone its differences on these metrics are not significant. The Bonferroni--Dunn test (critical difference 4.50) finds no method significantly better than FLIC on any metric (Supplementary Fig.~S2). With ML-kNN, FLIC is ahead on coverage, macro-AUC and Tail-AUC against all three embedded methods and on ranking loss against RFS and MIFS, significantly so against RFS and MIFS on macro-AUC and against MIFS on Tail-AUC; with logistic regression it is significantly better than all three on ranking loss and than RFS and MIFS also on coverage, macro-F1 and macro-AUC. The embedded methods, which aggregate the fitting errors of all labels in a utilitarian way, are ahead on Hamming loss, one-error and micro-F1 with both classifiers, and with ML-kNN also on average precision; in the Wilcoxon--Holm tests they are significantly better for RFS and LLSF on one-error and micro-F1 with ML-kNN, and for LLSF on both and RFS on micro-F1 with logistic regression. Macro-F1 is mixed (LLSF is ahead with ML-kNN, FLIC is first with logistic regression). On the largest data set, tmc2007-500 (28\,596 instances), FLIC has the best ranking loss, coverage and average precision of all 17 methods, whereas on bibtex (159 labels) RFS is best on every metric.

\subsection{Which labels benefit}\label{sec:labels}

\begin{figure}[!t]
\centering
\includegraphics[width=0.71\linewidth]{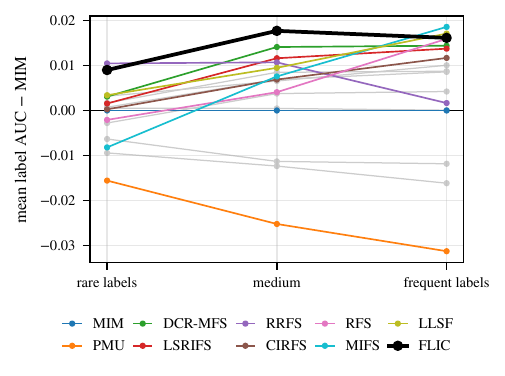}
\caption{Mean per-label AUC by label-frequency tercile, relative to MIM (averaged over subset sizes, folds and the 22 data sets). FLIC is best on medium-frequency labels and second on rare labels, without the loss on frequent labels that RRFS incurs. Thin light-grey lines: D2F, SCLS, MDMR, LRFS, STFS, GMM and ENM.}
\label{fig:tercile}
\end{figure}

Splitting the labels of each data set into terciles of their prior shows where FLIC's advantage comes from (Fig.~\ref{fig:tercile}). FLIC has the highest mean per-label AUC of all 17 methods on medium-frequency labels (0.705) and the second highest on rare labels (0.658, behind RRFS with 0.659), whereas MIFS and LLSF are best on frequent labels, followed by FLIC (0.734), the best of the 14 filters. RRFS, the most strongly fair baseline, pays for its rare-label AUC with a clearly lower AUC on frequent labels (0.720 against 0.734 for FLIC). FLIC raises the rare-label AUC on 14 to 20 of the 22 data sets against every baseline except RRFS, which is ahead on 14 of them. Relative to MIM, most of this advantage already comes from the label-wise, bias-corrected conditional gains: the utilitarian variant ($\alpha=0$) reaches 0.655 on rare and 0.702 on medium-frequency labels. Proportional fairness adds a small but consistent gain on top---0.0023 on rare and 0.0024 on medium-frequency labels, higher on 15 and 16 of the 22 data sets, respectively---with no loss on frequent labels. Against the strongest filters this gain turns ties into leads: on medium-frequency labels the utilitarian variant is ahead of DCR-MFS and LSRIFS on only 11 and 10 of the 22 data sets, whereas the default is ahead of each on 16.

\subsection{The fairness spectrum (Q2)}\label{sec:spectrum}

\begin{figure}[!t]
\centering
\includegraphics[width=\linewidth]{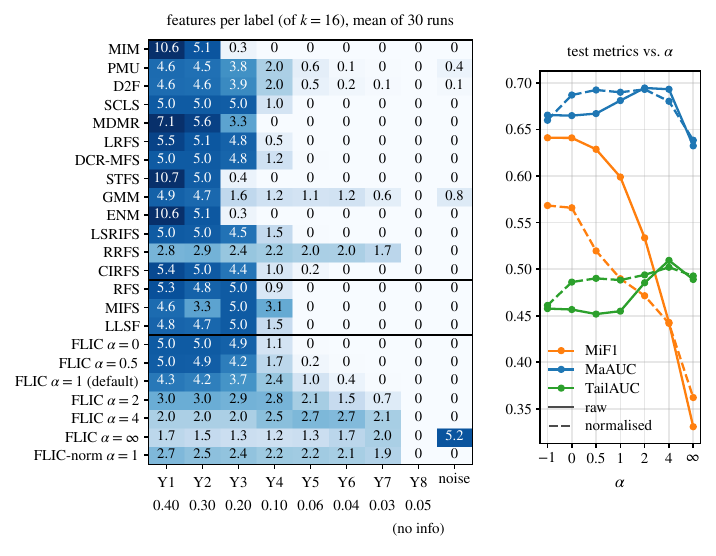}
\caption{Controlled study. Left: average number of the $k=16$ selected features that carry information about each label (priors under the label names) or about no label. Right: test metrics of FLIC as a function of $\alpha$ on the raw (solid) and normalised (dashed) coverage scales.}
\label{fig:synthetic}
\end{figure}

\paragraph{Controlled study}
We generated 30 data sets with eight labels of priors $0.40$, $0.30$, $0.20$, $0.10$, $0.06$, $0.04$, $0.03$ and $0.05$; each of the first seven labels owns five informative features, the two most frequent ones also own ten near-duplicates (a redundancy trap), the eighth has none, and 50 features are noise (Supplementary Section~S5). Every method selects $k=16$ features (Fig.~\ref{fig:synthetic}). MIM, STFS and ENM spend almost the whole budget (15.6--15.7 features) on the two most frequent labels, duplicates included. The redundancy-aware filters with little or no fairness (SCLS, LRFS, DCR-MFS, LSRIFS, CIRFS and FLIC with $\alpha\le0.5$) and the three embedded methods avoid many duplicates but still spend all, or almost all, of the budget (15.8--16.0 of 16 features) on the four most frequent labels (Fig.~\ref{fig:synthetic} and Supplementary Table~S13); all of them except MIFS obtain the lowest Hamming loss (0.089--0.092) and the highest micro-F1 (0.63--0.64) of the study. As $\alpha$ grows, FLIC spreads the budget and selects fewer duplicates: $\alpha=1$ covers 5.4 of the 7 informative labels on average, $\alpha=2$ covers 6.6 and maximises macro-AUC, and $\alpha=4$ covers all seven with an almost uniform allocation and maximises Tail-AUC. Macro-F1, in contrast, falls with $\alpha$ in this study (0.275 at $\alpha=0$, 0.263 at $\alpha=1$ and 0.212 at $\alpha=4$), unlike on the benchmarks. None of the compared methods offers such control over the allocation. The max--min limit spends a third of the budget (5.2 of 16 features) on noise, trying to raise the coverage of the label without informative features, as the leximin limit of Proposition~\ref{prop:maxmin} demands---the price of extreme fairness under estimation noise.

\begin{figure}[!t]
\centering
\includegraphics[width=\linewidth]{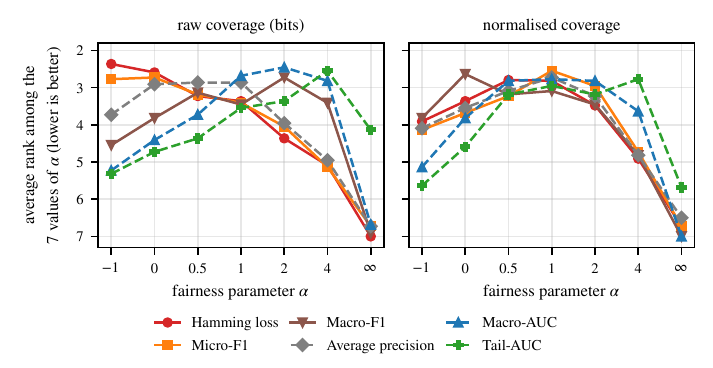}
\caption{Average rank of FLIC with seven values of $\alpha$ on six metrics (lower is better). On the raw scale, Hamming loss and micro-F1 prefer small $\alpha$ and the label-wise metrics (macro-F1, macro-AUC, Tail-AUC) moderate $\alpha$; the max--min limit is the worst on almost every metric.}
\label{fig:alpha}
\end{figure}

\paragraph{Benchmark data}
Over the 22 data sets, the pattern suggested by Theorem~\ref{thm:bounds} is clearly visible on the raw scale. Hamming loss and micro-F1 deteriorate steadily as $\alpha$ grows (Hamming loss: average rank 2.36 at $\alpha=-1$ and 5.09 at $\alpha=4$; micro-F1: 2.73 at $\alpha=0$ and 5.14 at $\alpha=4$), whereas the label-wise metrics improve up to a moderate value: macro-F1 and macro-AUC are best at $\alpha=2$ and Tail-AUC at $\alpha=4$, where its mean rises to 0.531 from 0.523 at $\alpha=-1$. Average precision is flat for $\alpha\in[0,1]$, ranking loss and coverage (not shown) are best at $\alpha=1$, and the max--min limit is the worst setting on every metric except Tail-AUC. On the normalised scale, Hamming loss, micro-F1, average precision and macro-AUC are best at $\alpha\in[0.5,1]$ and macro-F1 at $\alpha=0$, and at $\alpha=1$ normalisation raises Tail-AUC (0.532 against 0.527) but lowers micro-F1 (0.455 against 0.471), in line with the development-set finding that normalisation over-invests in rare labels under heavy imbalance. The variations within $\alpha\in[0,2]$ are modest (Fig.~\ref{fig:alpha}), so the default $\alpha=1$ is a safe compromise.

\subsection{Ablation and sensitivity (Q3)}\label{sec:ablation}

\begin{table}[!t]
\centering
\caption{Ablation study. First row: metric values of the default FLIC averaged over the 22 data sets. Other rows: number of data sets on which the default is better/worse than the variant that changes one component. Arrows indicate the preferred direction.}
\label{tab:ablation}
\footnotesize
\setlength{\tabcolsep}{3pt}
\begin{adjustbox}{max width=\linewidth}\begin{tabular}{lccccccccc}
\toprule
Variant & HL$\downarrow$ & RL$\downarrow$ & CV$\downarrow$ & AP$\uparrow$ & OE$\downarrow$ & MiF1$\uparrow$ & MaF1$\uparrow$ & MaAUC$\uparrow$ & TailAUC$\uparrow$\\
\midrule
FLIC (default), mean value & 0.0776 & 0.122 & 0.222 & 0.674 & 0.367 & 0.471 & 0.247 & 0.699 & 0.527\\
\midrule
mean-CMI gain (JMI-type) & 17/5 & 17/5 & 18/4 & 18/4 & 18/4 & 19/3 & 18/4 & 16/6 & 12/10\\
marginal MI gain & 18/4 & 20/2 & 20/2 & 20/2 & 20/2 & 19/3 & 19/3 & 18/4 & 13/9\\
no bias correction & 15/7 & 20/2 & 18/4 & 16/6 & 15/7 & 18/4 & 19/3 & 17/5 & 11/11\\
first-order gain & 12/10 & 10/12 & 11/11 & 9/13 & 11/11 & 8/14 & 7/15 & 13/9 & 12/10\\
normalised coverage & 20/2 & 18/4 & 16/6 & 18/4 & 18/4 & 19/3 & 16/6 & 15/7 & 4/18\\
utilitarian ($\alpha=0$) & 6/16 & 18/4 & 17/5 & 11/11 & 9/13 & 8/14 & 12/10 & 18/4 & 16/6\\
\bottomrule
\end{tabular}\end{adjustbox}
\end{table}

Table~\ref{tab:ablation} compares the default FLIC with variants that change one component at a time. Replacing the label-wise conditional gain of Eq.~(\ref{eq:cmim}) by the marginal mutual information, which reduces FLIC to a label-reweighted ranking without redundancy handling, causes the largest loss: the default is better on 18--20 of the 22 data sets for eight metrics. The variants that average the conditional information over the selected features (JMI-type) instead of taking the minimum, or that omit the bias correction, lose on 15--20 data sets for every metric except Tail-AUC, confirming that conservative, debiased estimates matter when the ledger accumulates them over many greedy steps. Using the first-order gain of Eq.~(\ref{eq:firstorder}) instead of the exact gain makes no systematic difference (the mean of every metric changes by less than 0.3\%). Normalised coverage (with $\varepsilon=0.1$) improves Tail-AUC (on 18 data sets) but loses on all other metrics, and the utilitarian setting $\alpha=0$ is better on Hamming loss, one-error and micro-F1 but worse on ranking loss, coverage, macro-AUC and Tail-AUC (on 16--18 data sets), as in Section~\ref{sec:spectrum}; for $\alpha=0$ the mean metric values differ from the default by at most 1\% (Supplementary Table~S17). The smoothing constant has a small effect: for $\varepsilon$ between $0.02$ and $1$ bit every metric stays, on average over the data sets, within $2\%$ of its value at the default (Supplementary Fig.~S4).

\subsection{Robustness to the classifier and to multi-view data (Q4)}\label{sec:robust}

Re-evaluating all rankings with binary-relevance logistic regression (Supplementary Section~S6) gives a similar picture, somewhat more favourable to FLIC: it has the best mean rank (4.64, against 5.79 for LLSF and 6.21 for DCR-MFS), is first on ranking loss, coverage, macro-F1 and macro-AUC and shares the first rank on average precision with LLSF, while LLSF is first on Hamming loss, one-error and micro-F1; 98 of the 144 pairwise comparisons are significant in favour of FLIC and three against it. Restricted to the 14 information-theoretic filters (Holm correction over the 13 filters, as in Section~\ref{sec:main}), FLIC is first on seven metrics, now including Hamming loss, shares the first rank on one-error with DCR-MFS and is second on Tail-AUC, behind SCLS; its mean rank is 3.20, against 4.48 for DCR-MFS, and 92 of the 117 comparisons are significant in its favour---on five to eight metrics against each of the 13 filters---and none against it. On five multi-view multi-label data sets with the features of all views concatenated (Supplementary Section~S7), FLIC has the best average rank on macro-AUC (2.6) and Tail-AUC (3.2) and, although it ignores the views, outperforms the view-aware filter RRFS on 34 of the 35 data-set/metric pairs and is on a par with the view-aware embedded method MSFS (19 wins, 16 losses); over the seven metrics reported there, LLSF has the best average rank (3.34), followed by DCR-MFS (4.11) and FLIC (4.31).

\section{Discussion}\label{sec:discussion}

\paragraph{Where the gains come from}
Measuring coverage \emph{per label} forces a per-label accounting of information---the ledger and the label-wise conditional gains---and the ablation identifies the conditional gains, together with bias correction, as the main source of accuracy (Table~\ref{tab:ablation}). Without its bias correction FLIC would rank second among the filters (mean rank 4.57, against 3.77 for DCR-MFS, though still first on coverage and macro-AUC), but the lead persists when all filters use the same correction (Section~\ref{sec:main}). Fairness trades Hamming loss, one-error and micro-F1 (worse on 13 to 16 of the 22 data sets) for ranking loss, coverage, macro-AUC and Tail-AUC (better on 16 to 18; Section~\ref{sec:ablation}). The fairness parameter $\alpha$ thus decides how the information is shared: it moves the selection along the trade-off suggested by Theorem~\ref{thm:bounds} and is best viewed as a statement of intent rather than a tuning knob. For Hamming loss or micro-F1, small $\alpha$ on the raw scale is preferable; when every label matters equally---rare diagnoses in medical coding, minority concepts in image annotation---moderate fairness ($\alpha$ between 1 and 4) lifts the worst-served labels at a small cost in average accuracy. Normalised coverage bounds each label's relative error (Theorem~\ref{thm:bounds}(b)), but with a handful of positive instances its estimate is noisy; under strong imbalance the raw scale with a smoothing $\varepsilon$ of a few tenths of a bit, which acts as a noise floor, is the safer default.

\paragraph{Relation to the strongest baselines}
DCR-MFS is a utilitarian filter with conditional relevance (Table~\ref{tab:taxonomy}); with ML-kNN it ranks better than FLIC only on Hamming loss, the metric that Theorem~\ref{thm:bounds} ties to utilitarian selection. LSRIFS lies in the concentration-seeking regime, lacks the guarantee of Theorem~\ref{thm:submod} and spends almost the whole budget on the four most frequent labels in the controlled study. The embedded methods LLSF and RFS are utilitarian with respect to the labels' fitting errors: they also serve only the four most frequent labels in the controlled study and lead on Hamming loss, one-error and micro-F1 (with ML-kNN also on average precision), possibly helped by their joint fit of the continuous feature values. Against all four, FLIC has the better mean rank on coverage, macro-AUC and Tail-AUC with both classifiers, needs no relevance--redundancy trade-off parameter and makes the allocation of the budget explicit and controllable.

\paragraph{Limitations}
FLIC inherits the limits of low-order information estimation: the label-wise CMIM surrogate ignores synergies among three or more features, and the plug-in estimator requires discretisation; neighbourhood estimators \cite{lin2016neighborhood} could replace it without changing the framework. The control with bias-corrected baselines was run with ML-kNN only. Our comparison with embedded methods is limited to three established linear models with one parameter setting chosen on the development data; newer embedded methods, such as those of Jia et al.\ \cite{jia2024discriminative} and Xu et al.\ \cite{xu2025orthogonal}, or per-data-set tuning may widen their advantage. Finally, the label weights $w_l$ were uniform throughout; prior knowledge of label importance can be injected through them.

\section{Conclusion}\label{sec:conclusion}

This paper started from a simple observation: whenever many labels share a small feature budget, a selection criterion decides which labels receive the information, and the criteria published so far make this decision implicitly and inconsistently. Casting multi-label feature selection as the $\alpha$-fair welfare of per-label information coverage turns the decision into one explicit parameter with a clear meaning, the marginal value of a bit of information to a label that is already well covered.

\paragraph{Strengths and benefits to the field}
The framework explains rather than adds to the list of heuristics: a dozen published relevance terms are located on one axis, so the apparent conflict between amplifying label-specific relevance and balancing it across labels reduces to a choice of $\alpha$. It also gives the $\alpha$-fair family guarantees that the existing heuristics do not state---submodularity and the $(1-1/e)$ greedy bound for every $\alpha\ge0$ when each label's information is submodular, their failure for concentration-seeking criteria, and the intractability of the max--min limit---and Fano and Hellman--Raviv bounds that indicate which $\alpha$ suits Hamming loss and which suits label-wise metrics. The algorithm derived from it, FLIC, has no relevance--redundancy trade-off parameter and was the strongest of the fourteen information-theoretic filters in our study: first on seven of nine metrics over 22 data sets with ML-kNN, first or joint first on eight with logistic regression, and never significantly worse than any filter with either classifier. Among all 17 methods, including three tuned embedded methods, it had the best mean rank, clearly with logistic regression and narrowly with ML-kNN, and the best rank on ranking loss, coverage and macro-AUC with both classifiers. It had the highest mean per-label AUC of the 17 methods on medium-frequency labels and the second highest on rare labels, 0.001 behind RRFS, while being the best of the fourteen filters on frequent labels, within 0.003 of the best embedded method. In a controlled study and, on the raw scale, on the benchmarks, increasing $\alpha$ to a moderate value ($2$ to $4$) traded Hamming loss and micro-F1 for macro-AUC and Tail-AUC, as the theory suggests.

\paragraph{Weaknesses}
The guarantees concern the true information functions under a conditional-independence condition, whereas FLIC optimises low-order, plug-in estimates computed after discretisation. This estimation step is the main practical limitation: extreme fairness amplifies the estimation noise of the rarest labels, which plausibly explains why the max--min limit performed worst, and the normalised scale, which the bounds tie to label-wise error, improved only Tail-AUC on the benchmarks. Fairness adds small but consistent gains on rare and medium-frequency labels and on ranking loss, coverage, macro-AUC and Tail-AUC at some cost in Hamming loss, one-error and micro-F1. With ML-kNN the margins over the strongest filters, DCR-MFS and LSRIFS, are mostly not significant, whereas with logistic regression FLIC is significantly better than each on five metrics. Tuned embedded methods remained ahead on Hamming loss, one-error and micro-F1 with both classifiers, and with ML-kNN also on average precision. Finally, the per-label view does not credit information carried only by combinations of labels.

\paragraph{Future work}
Neighbourhood, kernel or variational estimators of information would sharpen the coverage ledger, remove the discretisation step and allow a label-adaptive noise floor that makes the normalised scale as reliable as the raw one. The welfare view is not limited to filters: embedded and deep feature selectors could replace the unweighted aggregate of per-label errors by an $\alpha$-fair one, and in multi-view, streaming or partially labelled settings the agents could be views, label groups or tasks. Finally, the label weights offer a principled entry point for domain knowledge, such as priorities elicited from experts, and $\alpha$ itself could be chosen to meet a target on the worst-served labels.

\section*{Declarations}
\paragraph{Data availability}
Eleven of the evaluation data sets and the development set Corel5k are publicly available from the Mulan repository \cite{tsoumakas2011mulan}, and Stackex-chemistry, Ohsumed and EukaryotePseAAC from the KDIS multi-label repository (\url{https://www.uco.es/kdis/mllresources/}). The eleven Yahoo! collections, in the reduced-vocabulary version of Zhang and Zhou \cite{zhang2007mlknn}, are available from the author on request in the form used here. The five multi-view data sets are those used by Han et al.\ \cite{han2024rrfs}: the four image data sets were prepared by Zhang et al.\ \cite{zhang2020msfs} from public image collections, and Yeast-MV splits the yeast data into two views (Supplementary Section~S7). They were obtained from Han et al.\ and are available from them on request. The synthetic data of the controlled study are generated by the experiment scripts (seeds 0--29). The code of FLIC and of all baselines, the experiment scripts and the raw per-fold results are available from the author on request.

\paragraph{CRediT authorship contribution statement}
\textbf{Zhenzhi Tang:} Conceptualization, Methodology, Formal analysis, Software, Investigation, Visualization, Writing -- original draft, Writing -- review \& editing.

\paragraph{Declaration of competing interest}
The author declares that there are no known competing financial interests or personal relationships that could have appeared to influence the work reported in this paper.

\paragraph{Funding}
This research did not receive any specific grant from funding agencies in the public, commercial, or not-for-profit sectors.

\paragraph{Declaration of generative AI and AI-assisted technologies in the manuscript preparation process}
During the preparation of this work the author used Claude Code (Anthropic), running the Claude Opus 5.5 model, as an assistant in order to implement the experiments and to draft and edit the manuscript, with the author and the tool working together throughout. After using this tool, the author reviewed all code, results and text in full, edited the content as needed and takes full responsibility for the content of the publication.

\bibliographystyle{elsarticle-num}
\bibliography{refs}

\section*{Author biography}
\noindent\textbf{Zhenzhi Tang} received the B.S. degree in computer science from Jilin University in 2025 and is a graduate student at the Information Networking Institute, Carnegie Mellon University. Research interests: multi-label learning, feature selection.

\end{document}